\documentclass[conference]{IEEEtran}
\IEEEoverridecommandlockouts
\usepackage{cite}
\usepackage{amsmath,amssymb,amsfonts}
\usepackage{algorithm}
\usepackage{algorithmicx}
\usepackage{algpseudocode}
\usepackage{graphicx}
\usepackage{textcomp}
\usepackage{xcolor}
\usepackage{dsfont}
\usepackage{multirow}
\usepackage{booktabs}
\usepackage{array}

\def\BibTeX{{\rm B\kern-.05em{\sc i\kern-.025em b}\kern-.08em
    T\kern-.1667em\lower.7ex\hbox{E}\kern-.125emX}}

\usepackage[caption=false]{subfig}

\begin{document}

\title{Prompt What You Need: Enhancing Segmentation in Rainy Scenes with Anchor-based Prompting
\thanks{This research was supported in part by the National Natural Science Foundation of China (NSFC) under Grant 61906014 and 61976017.}

}

\author{
\IEEEauthorblockN{Xiaoyu Guo, Xiang Wei\thanks{Xiang Wei is the corresponding author.}, Qi Su, Huiqin Zhao and Shunli Zhang}
\IEEEauthorblockA{School of Software Engineering, Beijing Jiaotong University}
}

\maketitle

\begin{abstract}
Semantic segmentation in rainy scenes is a challenging task due to the complex environment, class distribution imbalance, and limited annotated data. To address these challenges, we propose a novel framework that utilizes semi-supervised learning and pre-trained segmentation foundation model to achieve superior performance. Specifically, our framework leverages the semi-supervised model as the basis for generating raw semantic segmentation results, while also serving as a guiding force to prompt pre-trained foundation model to compensate for knowledge gaps with entropy-based anchors. In addition, to minimize the impact of irrelevant segmentation masks generated by the pre-trained foundation model, we also propose a mask filtering and fusion mechanism that optimizes raw semantic segmentation results based on the principle of minimum risk. The proposed framework achieves superior segmentation performance on the Rainy WCity dataset and is awarded the first prize in the sub-track of STRAIN in ICME 2023 Grand Challenges. 
\end{abstract}


\section{Introduction}
Semantic segmentation is a critical task in computer vision that involves assigning a class label to each pixel in an image. Semantic segmentation has numerous applications, including autonomous driving~\cite{feng2020deep}, surveillance~\cite{pierard2023mixture}, and robotics~\cite{li2022towards}. Despite significant advancements in semantic segmentation, accurately segmenting images in complex scenarios, such as rainy scenes~\cite{zheng2020single,zhong2022rainy}, remains a formidable challenge. In general, rainy scenes introduce significant complexities due to factors such as environmental variability, class distribution imbalance, and the scarcity of annotated data. These challenges frequently result in a decline in segmentation performance, underscoring the need for enhanced methods capable of effectively addressing such specific scenarios.

\begin{figure}[ht]
  \centering
  \subfloat[Input Image]{
    \includegraphics[width=0.45\linewidth]{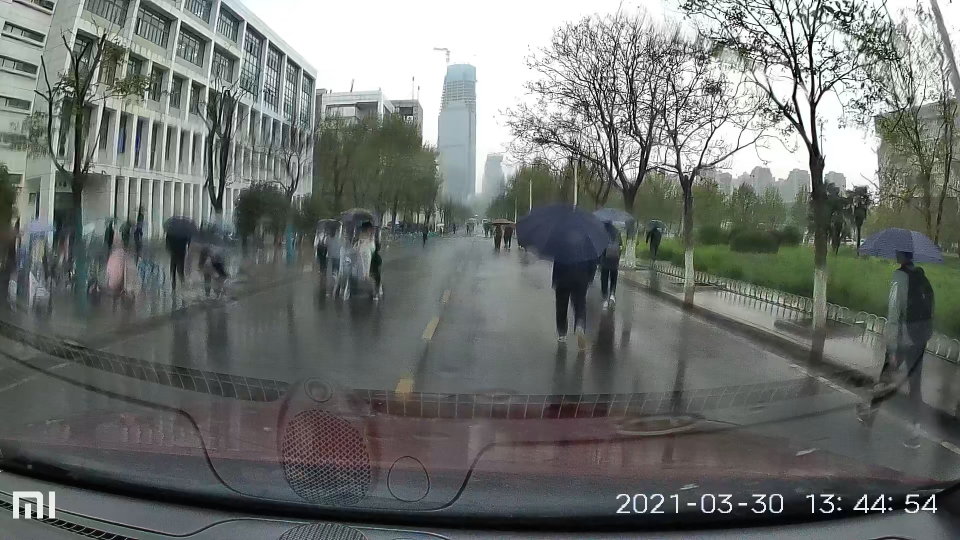}
    \label{fig:intro1}
  }
  \hfill
  \subfloat[Segment Anything Model]{
    \includegraphics[width=0.45\linewidth]{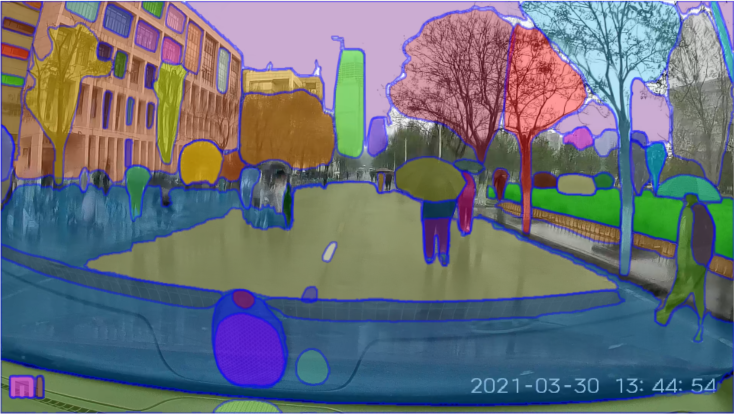}
    \label{fig:intro2}
  }
  \caption{Demonstration of SAM. In sub-figure (b), the performance of SAM is influenced by environmental factors, particularly reflections in rainy scenes. Additionally, the segmentation results generated by SAM may not fully adapt to specific tasks and may also potentially ignore small entities.}
  \label{fig:intro}
\end{figure}

Recent developments in foundation segmentation models, such as Segment Anything Model (SAM)~\cite{kirillov2023segany} and SegGPT~\cite{wang2023seggpt}, have demonstrated impressive results on a wide range of segmentation tasks in zero-shot scenarios. However, these models do not perform up to expectations when it comes to semantic segmentation in rainy scenes (as shown in Fig. \ref{fig:intro}). These limitations prompt the necessity for a novel framework that can harness the power of pre-trained foundation models without retraining while addressing the unique challenges posed by rainy scenes.

In this paper, we devise an innovative framework that combines semi-supervised techniques with pre-trained foundation models to effectively tackle semantic segmentation in rainy scenes with limited labeled training data. In brief, our framework consists of the following three steps:
\begin{itemize}
    \item We first leverage semi-supervised base model U$^2$PL\cite{wang2022semi} to provide guidance information, as it is capable of utilizing unreliable pixels for representation learning by contrastive learning. This is suitable for handling uncertainty caused by environment interference.
    \item We further propose to use the high-entropy regions calculated from U$^2$PL's predictions to generate anchors for prompting SAM. This strategy enables the identification of entities heavily impacted by rainy scenes, which tend to be more challenging to classify accurately.
    \item Finally, we put forward a filtering and fusion mechanism that carefully utilizes the segmentation masks generated by SAM to refine the predictions made by U$^2$PL. 

\end{itemize}

The proposed framework achieves superior segmentation performance on the Rainy WCity dataset and is awarded the first prize in the Seeing Through the Rain (STRAIN) -- Track 1: Semantic Segmentation under Real Rain Scene, which is part of Grand Challenges in the International Conference of Multimedia and Expo 2023. Furthermore, our framework also provides insights and inspiration for active prompting in promptable foundation models.

\section{Related Work}

Real-world semantic segmentation in rainy scenes presents several challenges. To solve the challenges, we provide a brief overview of semi-supervised learning in semantic segmentation and foundation models for computer vision.

\subsection{Semi-Supervised Semantic Segmentation}

Semi-supervised learning\cite{sohn2020fixmatch,kong20223lpr} has been widely adopted to overcome the challenge of limited labeled data. Early methods such as those proposed in \cite{mittal2019semi} leverage generative adversarial networks to train on unlabeled data using an adversarial loss, thereby reducing the gap between predictions on labeled and unlabeled data. In recent years, consistency regularization\cite{ouali2020semi}, self-training\cite{he2021re, yang2022st++}, and their combinations\cite{sohn2020fixmatch, zou2020pseudoseg, wei2021fmixcutmatch, chen2021semi} have become the mainstream in semi-supervised semantic segmentation. These methods aim to use unlabeled data to improve model performance while reducing the impact of label noise, such as weak and strong augmentations of the same sample. Improving the quality of trusted pseudo-labels is crucial for self-training and enhancing model performance \cite{chen2021semisupervised}. Additionally, contrastive learning has shown promising results in semi-supervised feature extraction\cite{alonso2021semi, zhong2021pixel}. U$^2$PL\cite{wang2022semi} uses the value of entropy for filtering reliable pixel-wise pseudo-labels and pushes the remaining unreliable pixels to a category-wise memory bank for contrastive learning, resulting in improved segmentation performance.

Unlike existing semi-supervised learning approaches, our framework exclusively relies on the semi-supervised model as the base model. The base model guides a pre-trained segmentation foundation model to bridge the knowledge gaps between the two models, leading to improved accuracy of the segmentation outcomes.

\subsection{Foundation Model for Computer Vision}

The field of natural language processing (NLP) is being revolutionized by large language models pre-trained on web-scale datasets, e.g., ChatGPT. These models, commonly referred to as "foundation models"~\cite{bommasani2021opportunities}, are capable of strong zero-shot and few-shot generalization, extending their capabilities to tasks and data distributions beyond those seen during training. Similarly, beyond NLP, foundation models in the field of computer vision are also becoming increasingly popular. CLIP~\cite{radford2021learning} and ALIGN~\cite{jia2021scaling} are examples of foundation models that adopted contrastive learning to train text and image encoders, enabling zero-shot generalization to novel visual concepts and data distributions using text prompts. While much progress has been made in vision and language encoders, many computer vision problems lack abundant training data. Recently, SAM~\cite{kirillov2023segany} and SegGPT~\cite{wang2023seggpt} are proposed for image segmentation, both SAM and SegGPT are promptable model and have been pre-trained on a broad dataset using a task that enables powerful generalization. Nevertheless, both SAM and SegGPT require manual examples to achieve the expected results and struggle to maintain good performance in semantic segmentation for rainy scenes, as illustrated in Fig. \ref{fig:intro2}.

Different from existing prompt methods for pre-trained segmentation foundation models, we leverage the uncertainty regions in the predictions of the semi-supervised base model to generate anchors. These anchors accurately identify the weaknesses of the semi-supervised base model and enable more effective utilization of the pre-trained foundation model's knowledge. Specifically, we use SAM in our framework.

\section{Methodology}

In this section, we detail our three-stage framework for semantic segmentation in rainy scenes.

\subsection{Overall Architecture}

\begin{figure*}
    \centering
    \includegraphics[width=1.0\linewidth]{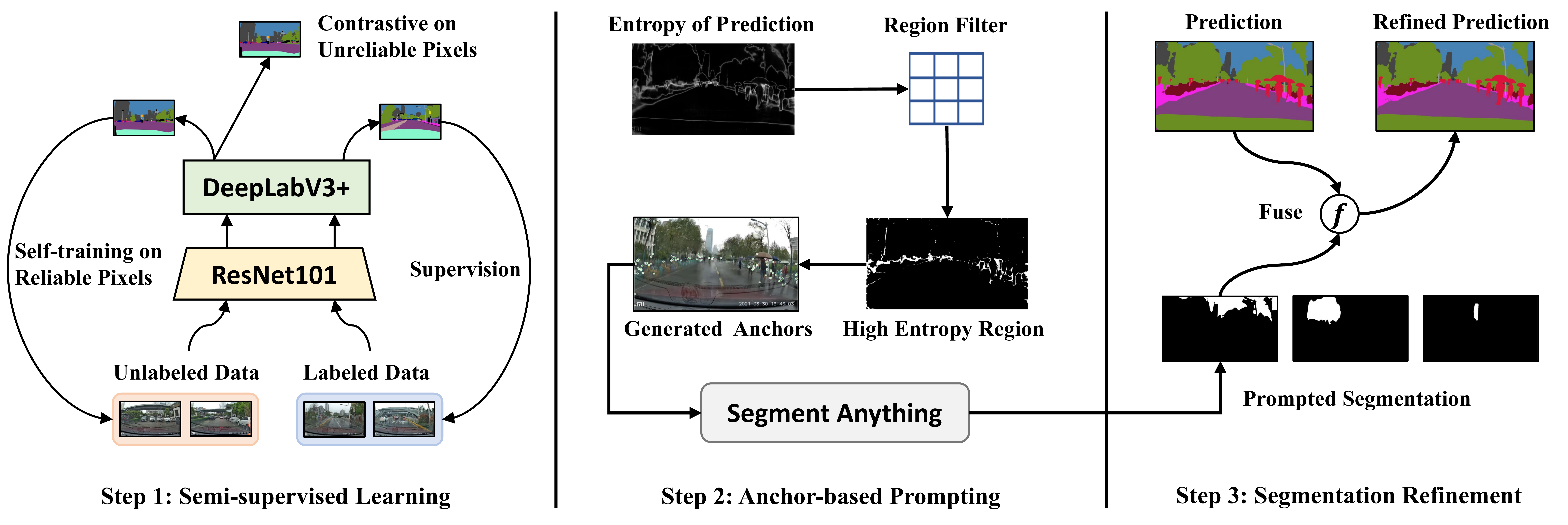}
    \caption{The overall process of our framework. The framework consists of three main steps. In step 1, we leverage semi-supervised learning to train a base model. In step 2, based on the predictions of the semi-supervised model, we calculate the entropy values of the predictions and remove the impact of segmentation boundaries to generate high-entropy regions, which are then treated as anchors. SAM generates semantic segmentation masks according to the generated anchors. In step 3, we optimize the predictions of the base model using the masks generated by SAM.}
    \label{fig:overall}
\end{figure*}

As depicted in Figure \ref{fig:overall}, our framework comprises three main steps. First, to overcome the challenge of limited annotated data, we employ semi-supervised learning to train the base model for semantic segmentation. Next, given the difficulty in establishing accurate semantics in areas affected by rainy environments in the existing dataset, we identify the image regions affected by interference by computing the entropy values of the semi-supervised base model's predictions. We then generate anchors in these image regions and use them to prompt the SAM to make predictions, resulting in a set of predicted segmentation masks. Finally, we meticulously leverage the segmentation masks generated by SAM to refine the original predictions.

\subsection{Semi-supervised Base Model Learning}

Since only one out of five of the training data has annotated labels, we suggest using a semi-supervised semantic segmentation model for initial training. For semi-supervised semantic segmentation, generating pixel-wise pseudo-labels in rainy scenes can be highly uncertain due to the environmental interference.  In the semi-supervised learning step, we follow the U$^2$PL~\cite{wang2022semi} and make suitable settings to it to adapt to semantic segmentation in rainy scenes. For annotated data, we adapt OHEM~\cite{shrivastava2016training}, which is responsible for mining difficult pixels and forces the model to focus on image regions that are affected by environmental interference, and this manner is effective for semantic segmentation in rainy scenes.

\subsection{Anchor-based Prompting}

\begin{figure}[ht]
  \centering
  \subfloat[Original Input]{
    \includegraphics[width=0.45\linewidth]{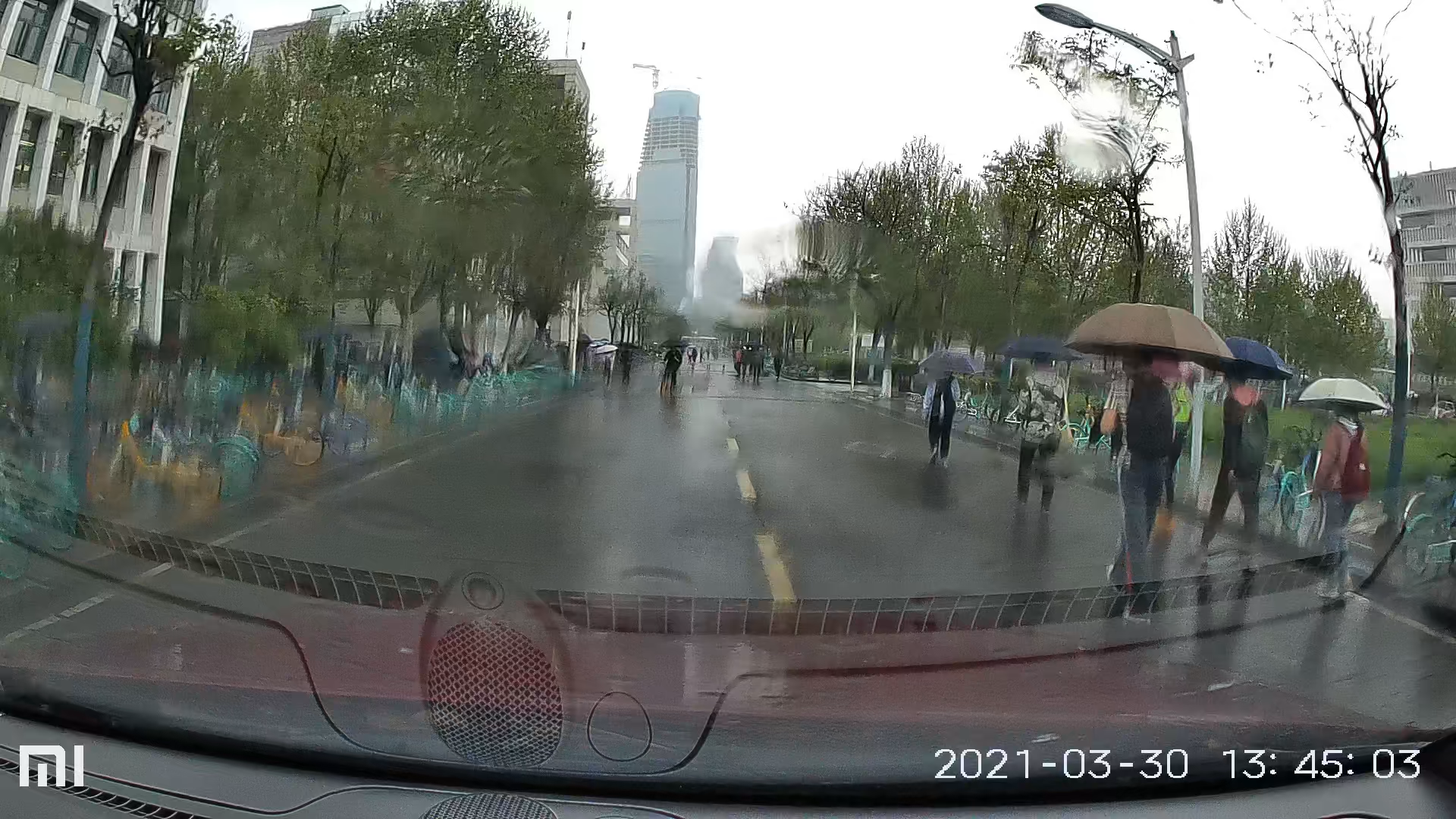}
    \label{fig:region2}
  }
  \subfloat[Entropy]{
    \includegraphics[width=0.45\linewidth]{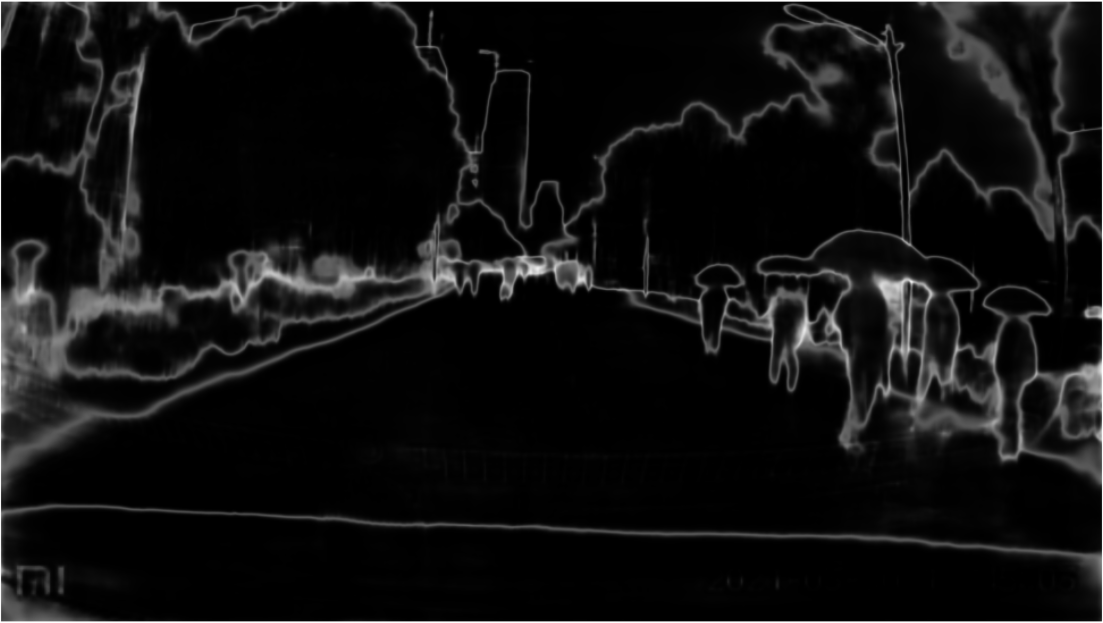}
    \label{fig:region1}
  }
  \hfill
  \subfloat[High-entropy Region]{
    \includegraphics[width=0.45\linewidth]{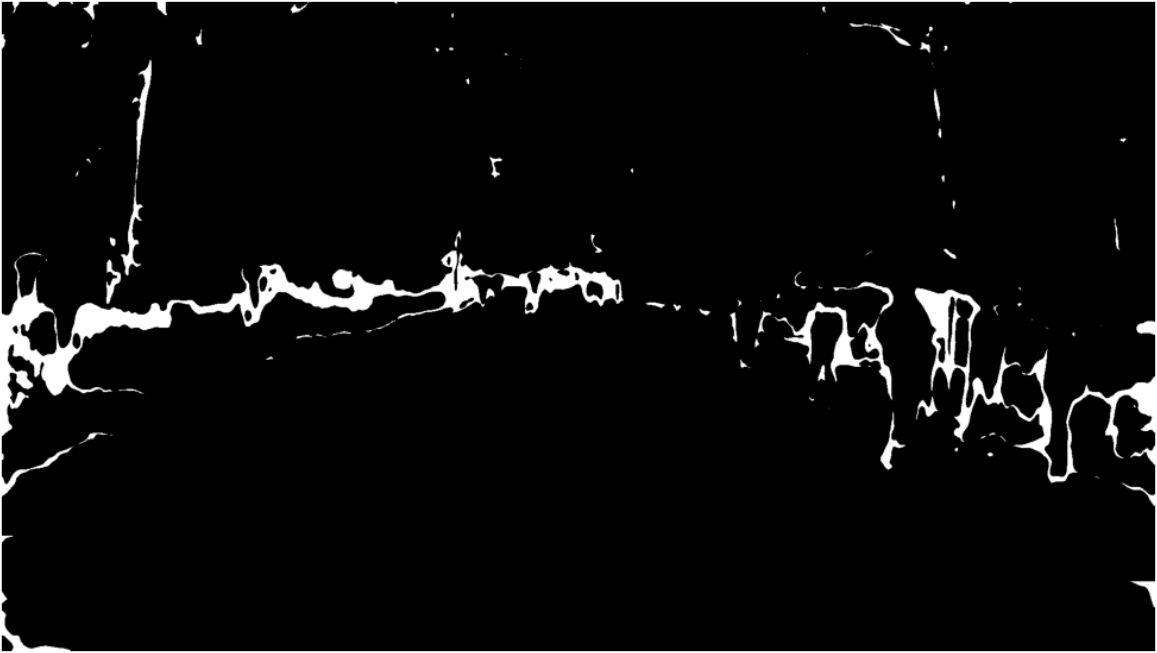}
    \label{fig:region2}
  }
  \subfloat[Anchors]{
    \includegraphics[width=0.45\linewidth]{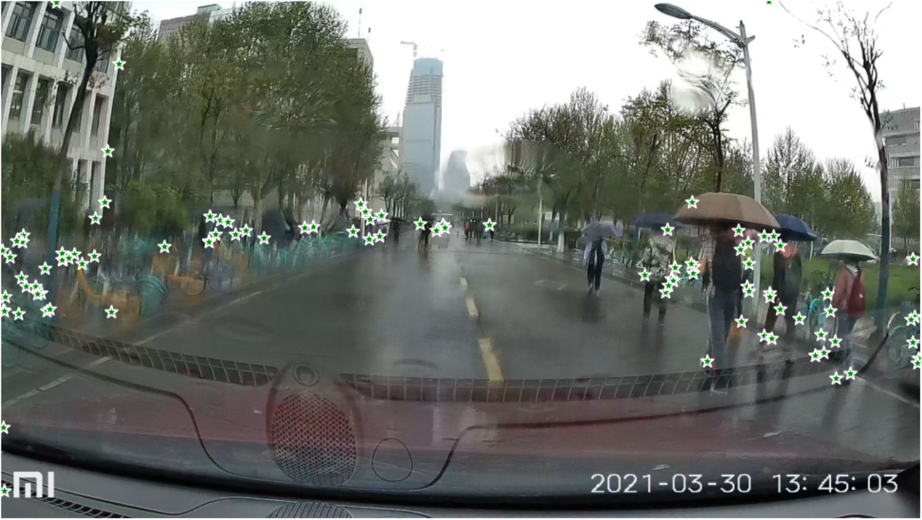}
    \label{fig:anchor}
  }
  \caption{The procedure of entropy-based anchor generation. Sub-figure (a) displays the original input image, while sub-figure (b) shows the entropy map produced by the semi-supervised base model. Sub-figure (c) highlights the regions of the image with high uncertainty after filtering using region filter and sub-figure (d) displays the sampled anchors marked with stars. }
  \label{fig:region}
\end{figure}

Considering that SAM can generate more accurate semantic segmentation results leveraging coordinate points as guidance. Furthermore, from our attempts, we have discovered that identifying the coordinate points according to the weakness of semi-supervised base model is crucial while using SAM to compensate for knowledge gaps. To pinpoint the model's weakness, we utilize the entropy values of predictions, which can be formalized by Equation~\eqref{eq:eq1}:

\begin{equation}\label{eq:eq1}
    {Ent}_{ij} = - \sum_{k=1}^{N} p_{ijk} \log p_{ijk}
\end{equation}
where ${N}$ denotes the number of classes, and ${p}_{ijk}$ stands for the ${k}$-th softmax score of the pixel in the ${i}$-th row and ${j}$-th column. By examining the entropy map displayed in Fig.~\ref{fig:region1}, we can discern that the boundaries of segmentation possess larger entropy values. The high entropy distribution caused by segmentation boundaries makes it difficult for us to pinpoint the image regions where the model is truly confused. 

To reduce the interference caused by segmentation boundaries, we design a region filter, which can also be interpreted as a $w \times w$ 2-D kernel with a fixed value 1 of weights then attached with a binarized activator. Besides, $w$ is an odd number to guarantee a explicit center. Equation~\eqref{eq:2} details the operations of the region filter:

\begin{equation}\label{eq:2}
    {Reg}_{ij}=f(Ent_{ij})=\mathds{1}_{\tau}(\frac{1}{w^2}\sum_{m=-w}^{w}\sum_{n=-w}^{w}{Ent}_{i+m,j+n}).
\end{equation}
Here, the function $f(\cdot)$ serves as a region filter, which converts the entropy map into a 0-1 mask with 1 indicating the high-entropy regions. ${Reg}_{ij}=f({Ent}_{ij})$ is the entropy value of the $i$-th row and $j$-th column pixel after being filtered and binarized. The binarized activator function with a threshold of $\tau$ is denoted by $\mathds{1}_{\tau}(\cdot)$. If the input is greater than or equal to $\tau$, $\mathds{1}_{\tau}(\cdot)$ returns 1; otherwise, it returns 0. The term $1/w^2$ is used for normalization. 

As illustrated in Fig.~\ref{fig:region1} and~\ref{fig:region2}, the filtered entropy map includes almost no finer segmentation boundaries, which only caused by intermediate pixels between different categories. Specifically, Fig.~\ref{fig:region1} displays the entropy map derived from the semi-supervised base model. As shown in Fig.~\ref{fig:region2}, undergoing the region filter, entropy values previously displayed by finer segmentation boundaries are no longer presented. On the contrary, regional high entropy values are retained with difficulty in classification caused by environmental interference and imbalanced class distribution. 

Based on the acquired high-entropy regions, we randomly sample coordinate points as anchors for prompting SAM to compensate for the lack of knowledge in the base model. The anchors generated by our method revealed in  Fig.~\ref{fig:anchor}. Then, we obtain corresponding segmentation results from SAM, which are binary masks for segmented entities.

\subsection{Segmentation Refinement}
Once the segmentation masks have been obtained from the SAM, they are supplemented to refine the predictions made by the semi-supervised base model. However, it is important to note that not all segmentation masks generated by the SAM from the anchors are reliable, so the key issue is how to extract useful knowledge from numerous masks to refine the predictions of the semi-supervised base model. Namely, we want to minimize the risk of reducing the accuracy of the existing segmentation results made by the semi-supervised base model when using the supplementary masks.

\textbf{\emph{The principle of minimum risk:}} Go back to Fig.~\ref{fig:intro2}, the segmentation masks generated by SAM contain a lot of interference information, such as incorrect entities and errors caused by environmental factors. To improve the quality of the masks during fusion, we propose to use two hyper-parameters to filter the masks. One is the softmax score $\alpha$, which filters low confidence segmentation masks; the other is a considerably small area $\beta$, which ensures the entity segmented by SAM is part of the target category, e.g., in Fig.~\ref{fig:intro2}, the "umbrella", "pants", and "shirt" are entities that segmented by SAM, but they belong to the "person" category. Furthermore, in case of a conflict, the priority of smaller segmented entities should be higher than that of larger entities. For instance, SAM may segment a "person" and a "bicycle" as a single entity based on some anchors, and we can correct this error by using the "person" entities segmented based on other anchors.

\begin{algorithm}[htbp]
    \caption{Segmentation Enhancement}
    \label{alg:1}
    \begin{algorithmic}[1]
        \Require Prediction $y$ and softmax score $p$, kernel size $w$, threshold of binarization $\tau$, threshold of softmax score $\alpha$, and threshold of area $\beta$
        \Ensure Enhanced prediction $y$
        \Function{SegEnhance}{$y$, $p$, $w$, $\tau$, $\alpha$, $\beta$}
        \State $Ent \gets$ computeEntropy($p$) \Comment{Eq.\ref{eq:eq1}}
        \State $Reg \gets$ getHighEntropyRegions($Ent$, $\tau$, $w$) \Comment{Eq.\ref{eq:2}}
        \State $Anc \gets$ generateAnchors($Reg$)
        \For{${Anc}_i$ in $Anc$}
            \State $M_i \gets$ segmentBySAM(${Anc}_i$)
        \EndFor
        \State $M   \gets$ filterByScoreAndArea($M$, $\alpha$, $\beta$) 
        \For{$M_i$ in $M$}
            \State $cls \gets$ getModeOfIntersaction($M_i$, $y$)
            \State Assign $cls$ to $M_i$
        \EndFor
        \State $M \gets$ sortByAreaFromHighToLow($M$) 
        \For{$M_i$ in $M$}
            \State Overwrite $y$ according to $M_i$ with class $cls$
        \EndFor
        \State \Return Enhanced prediction $y$
        \EndFunction
    \end{algorithmic}
\end{algorithm}

Algorithm 1 outlines the overall process of our framework. The algorithm mainly takes two inputs, namely the prediction $y$ and the corresponding softmax score $p$ from the semi-supervised base model. Additionally, four hyper-parameters are provided: $w$ represents the kernel size, and $\tau$ represents the binarization threshold in Equation~\eqref{eq:2}. Furthermore, $\alpha$ and $\beta$ denote the thresholds for the softmax score and area, respectively, which are utilized to ensure the quality of the segmentation masks produced by the SAM.

To be specific, in Algorithm~\ref{alg:1}, we first calculate the entropy values based on the softmax score $p$ by "computeEntropy" according to Equation~\eqref{eq:eq1}. Next, we call "getHighEntropyRegions" to apply a region filter based on Equation~\eqref{eq:2}, using hyperparameters $w$ and $\tau$, to eliminate the influence of segmentation boundaries and retain high-entropy regions. After obtaining the high-entropy regions, we adopt "generateAnchors" to perform sampling and generate anchor points. With the help of anchors to prompt SAM, we then obtain the corresponding segmentation masks by "segmentBySAM". Once we obtain the segmentation masks $M$ generated by SAM, we retain the low risk masks $M$ by their softmax score and area using the function "filterByScoreAndArea", specifically,  we aim to obtain entities that are segmented by SAM and have small area and high confidence. Then, for each mask $M_i$ in $M$, we find the mode of intersection between $M_i$ and $y$ by the function "getModeOfIntersection", which serves as the class for refining $y$ according to $M_i$. Next, we sort the remaining masks $M$ by their area from high to low using the function "sortByAreaFromHighToLow", this ensures that the results of smaller entities are not covered. At last, we let $M_i$ to determine the region that needs to be overwritten in prediction $y$, and write the corresponding class $cls$ to that region.

\begin{table*}[htbp]
\centering
\selectfont
\setlength{\tabcolsep}{0.45em}
\setlength{\extrarowheight}{2pt}
\caption{The top 5 teams' evaluation results for the Seeing Through the Rain: Track 1 - Semantic Segmentation under Real Rain Scene, which is part of the Grand Challenge Proposals of the IEEE International Conference on Multimedia and Expo 2023}
\label{tab:1}
\begin{tabular}{cccccccccccccccccccc}
\toprule
\textbf{Method} & \rotatebox{90}{road} & \rotatebox{90}{sidew.} & \rotatebox{90}{build} & \rotatebox{90}{wall} & \rotatebox{90}{fence} & \rotatebox{90}{pole} & \rotatebox{90}{light} & \rotatebox{90}{sign} & \rotatebox{90}{veget.} & \rotatebox{90}{sky} & \rotatebox{90}{person} & \rotatebox{90}{rider} & \rotatebox{90}{car} & \rotatebox{90}{truck} & \rotatebox{90}{bus} & \rotatebox{90}{motorc.} & \rotatebox{90}{bicycle} & \textbf{mIoU}\\
\hline
Rank 5 & 87.08 & 38.99 & 65.55 & 69.92 & 28.08 & 16.52 & 6.30  & 35.55 & 73.83 & 85.83 & 28.99 & 3.04 & 64.24 & 4.57 & 18.07 & 6.60 & 26.44 & 38.80\\
Rank 4 & 86.29 & 14.46 & 72.63 & 69.72 & 69.36 & 19.82 & 14.25 & 43.89 & 65.62 & 95.14 & 37.28 & 8.34 & 69.17 & 2.66 & 20.56 & 6.48 & 33.81 & 42.91\\
Rank 3 & 84.94 & 24.04 & 75.90 & 47.01 & 68.22 & 27.66 & 40.50 & 51.25 & 83.82 & 94.67 & 41.76 & 0.17 & 78.89 & 0.00 & 70.27 & 27.92 & 36.06 & 50.18\\
Rank 2 & 94.62 & \underline{61.13} & 84.76 & 84.19 & \underline{78.94} & 39.24 & 58.68 & \underline{80.36} & 86.46 & 96.36 & 24.95 & \underline{18.49} & 85.26 & \underline{9.10} & 70.33 & \underline{29.86} & 48.78 & 61.85\\
\textbf{Rank 1 (Ours)}  & \underline{94.64} & 53.88 & \underline{85.94} & \underline{87.05} & 78.52 & \underline{46.86} & \underline{62.96} & 79.58 & \underline{88.75} & \underline{96.51} & \underline{54.83} & 6.63 & \underline{87.14} & 7.77 & \underline{83.64} & 26.72 & \underline{50.69} & \textbf{64.24}\\
\bottomrule
\end{tabular}
\end{table*}

\section{Experiments}

In this section, we report the experimental results of the proposed framework for semantic segmentation in rainy scenes on the Rainy WCity dataset.

\subsection{Experimental Setup}
\textbf{Datasets}: We conduct our experiments on the Rainy WCity dataset, which consists of 500 images for training and 100 images for evaluation, all with a resolution of $1920\times1080$. Out of the 500 images, only 100 have pixel-wise annotations, while the remaining images are unannotated. Specifically, there are 240 raindrop images, 40 of which are annotated, 130 reflection images with 30 annotated, and 130 wiper images with 30 annotated. The dataset includes pixel-level labels for a total of 18 classes, including the background.

\textbf{Implementation details:} For semi-supervised base model (U$^2$PL), the experiment runs for 200 epochs, and we choose the checkpoint of the last epoch for evaluation. The network is based on the ResNet101 and Deeplabv3+ for encoder and decoder, respectively. We dynamically drop 20\% to 0\% of high-entropy pixels due to the unreliability while self-training. During training, we set the batch size to 2 for each of GPU. For anchor-based prompting, we set $w$ to 5 and $\tau$ to 1.0. We sample 1,000 anchors for each of prediction from the semi-supervised base model. For segmentation refinement, we set the threshold of softmax score $\alpha$ to 0.7 and threshold of area $\beta$ to 20,000, respectively. The experiments were conducted using Pytorch 1.12.0 and the entire training process was completed on 8 NVIDIA 3090 GPUs.

\begin{figure}
\centering
  \subfloat[Without Filtering]{
    \includegraphics[width=0.9\linewidth]{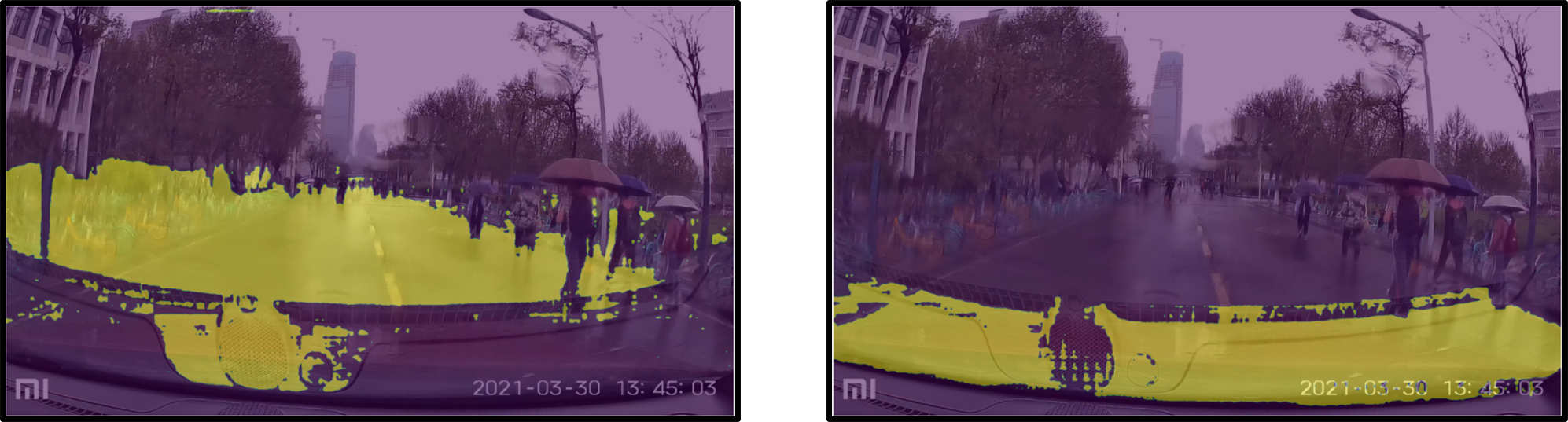}
    \label{fig:mask2}
  }
  \hfill
  \subfloat[Filtered by Softmax Score and Area]{
    \includegraphics[width=0.9\linewidth]{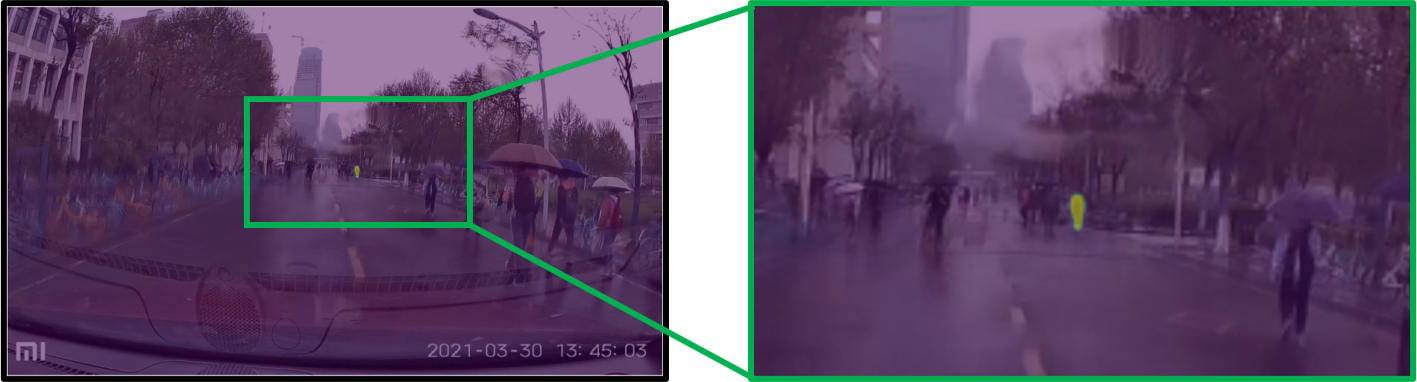}
    \label{fig:mask1}
  }
\caption{Segmentation by anchor-based prompting. The yellow region represents segmented entities by SAM. Sub-figure (a) shows the entities segmented without filtering, while sub-figure (b) shows the entities filtered based on both softmax score and area. The filtering function "filterByScoreAndArea" can be find on line 8 of Algorithm~\ref{alg:1}.}
\label{fig:mask}
\end{figure}

\subsection{Results}
\begin{figure*}[t]
\centering
\subfloat{\includegraphics[width=0.16\linewidth]{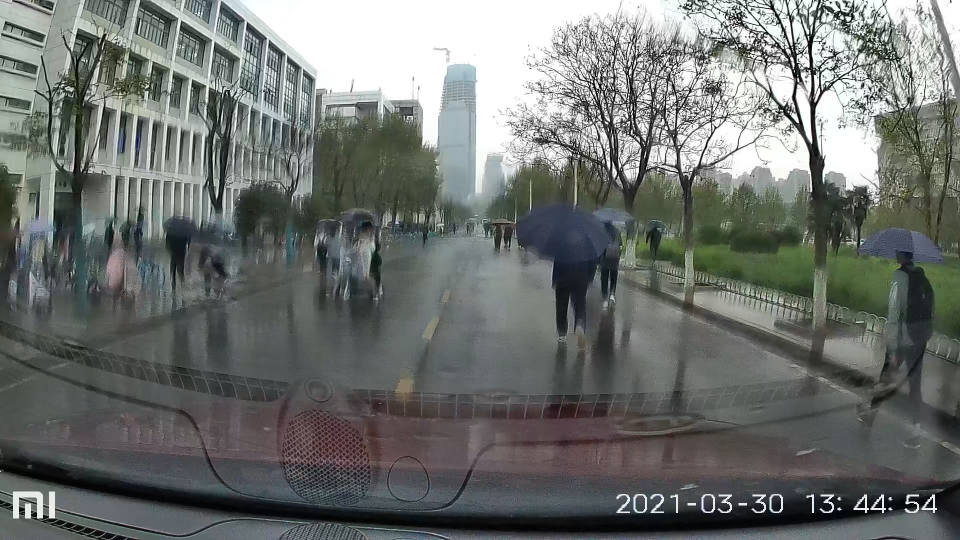}}
\hfill
\subfloat{\includegraphics[width=0.16\linewidth]{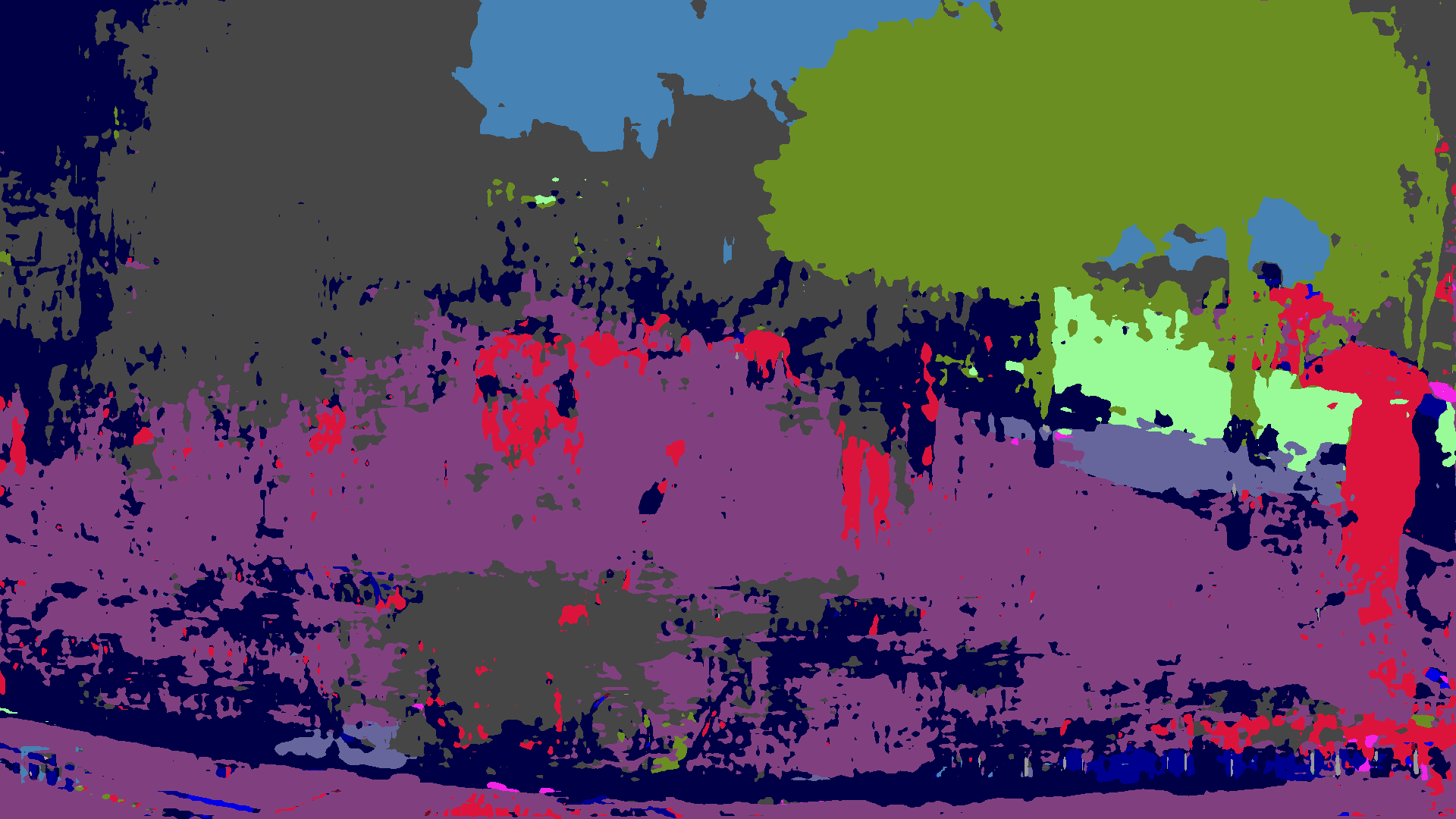}}
\hfill
\subfloat{\includegraphics[width=0.16\linewidth]{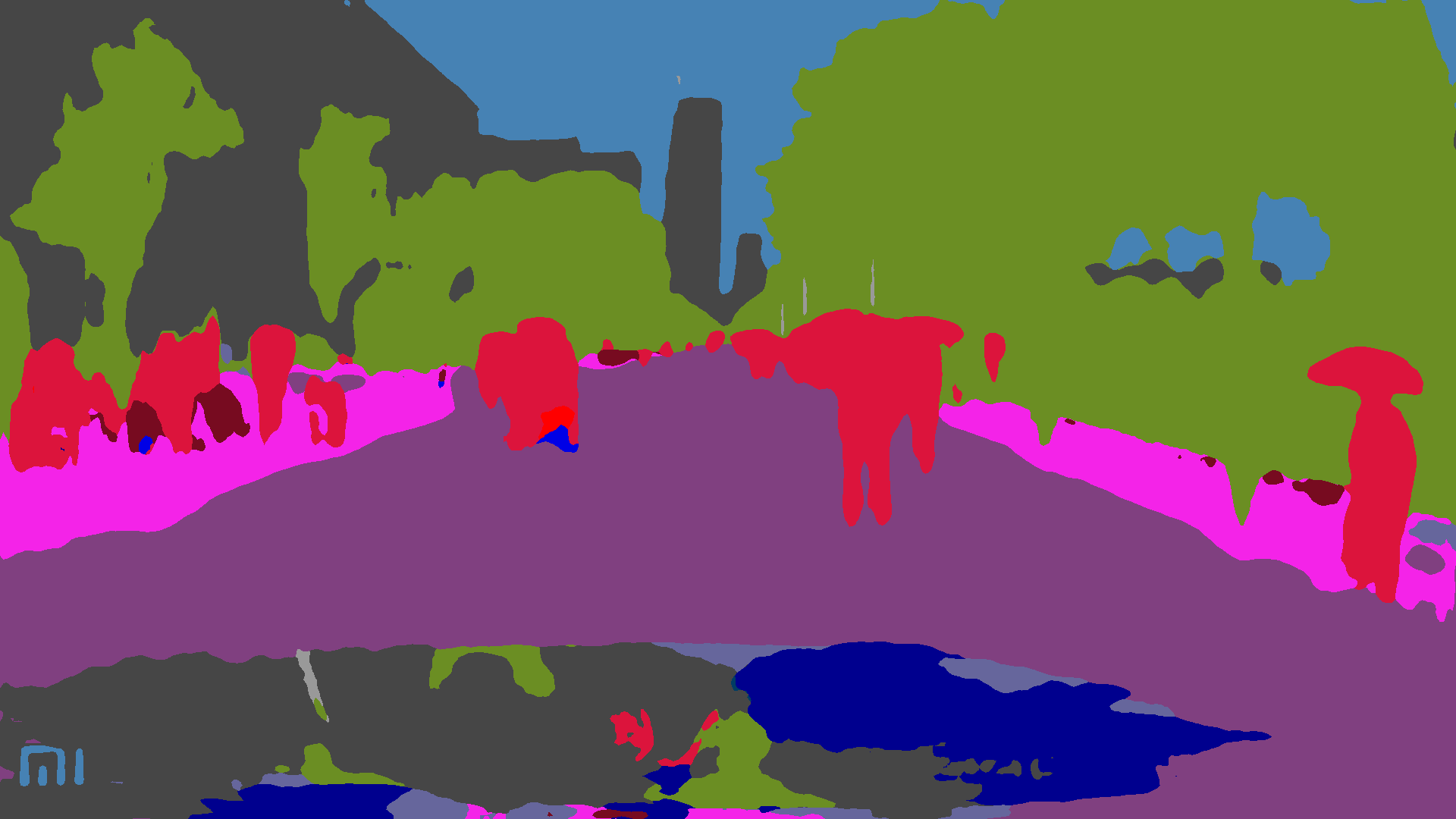}}
\hfill
\subfloat{\includegraphics[width=0.16\linewidth]{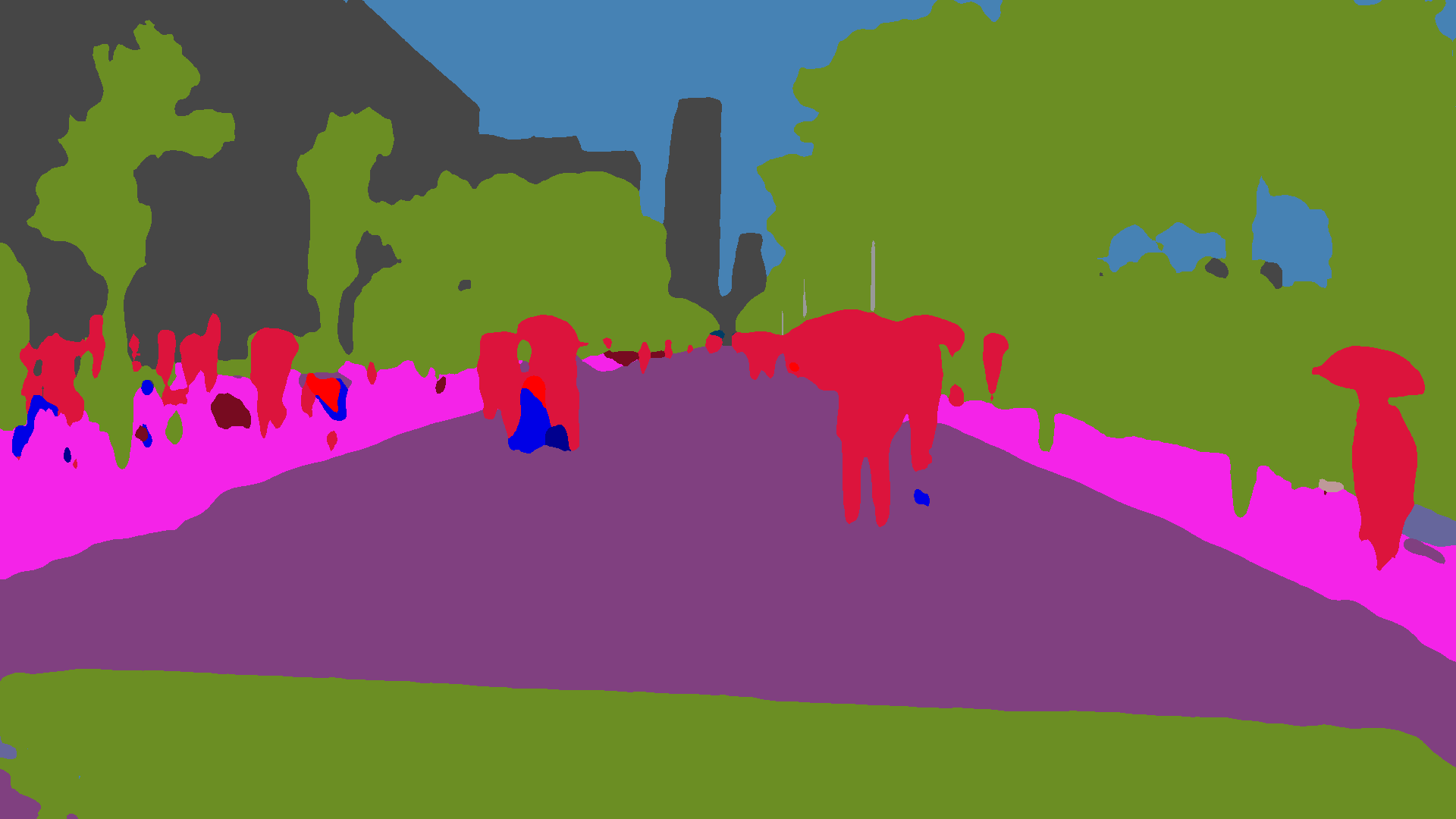}}
\hfill
\subfloat{\includegraphics[width=0.16\linewidth]{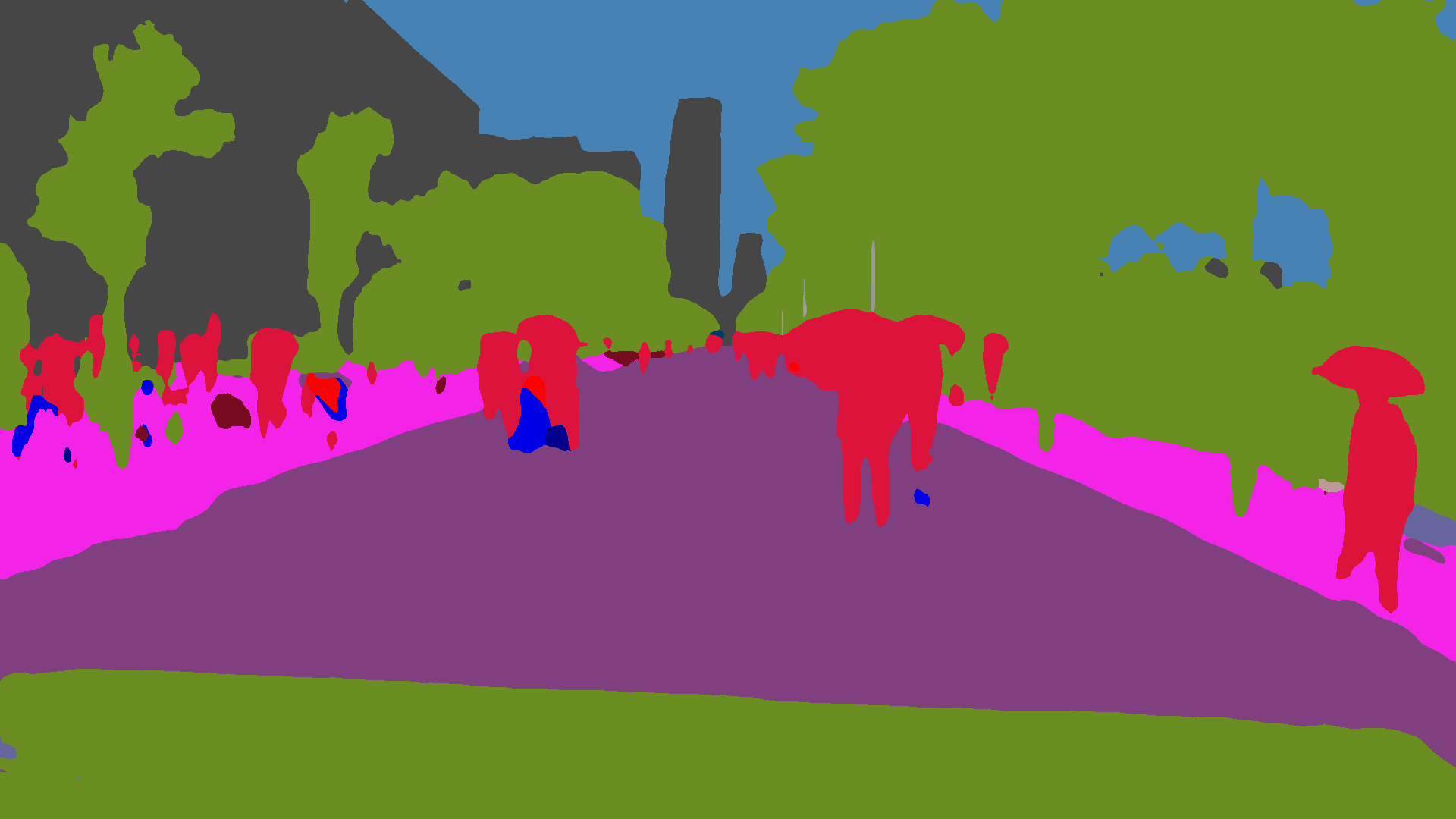}}
\hfill
\subfloat{\includegraphics[width=0.16\linewidth]{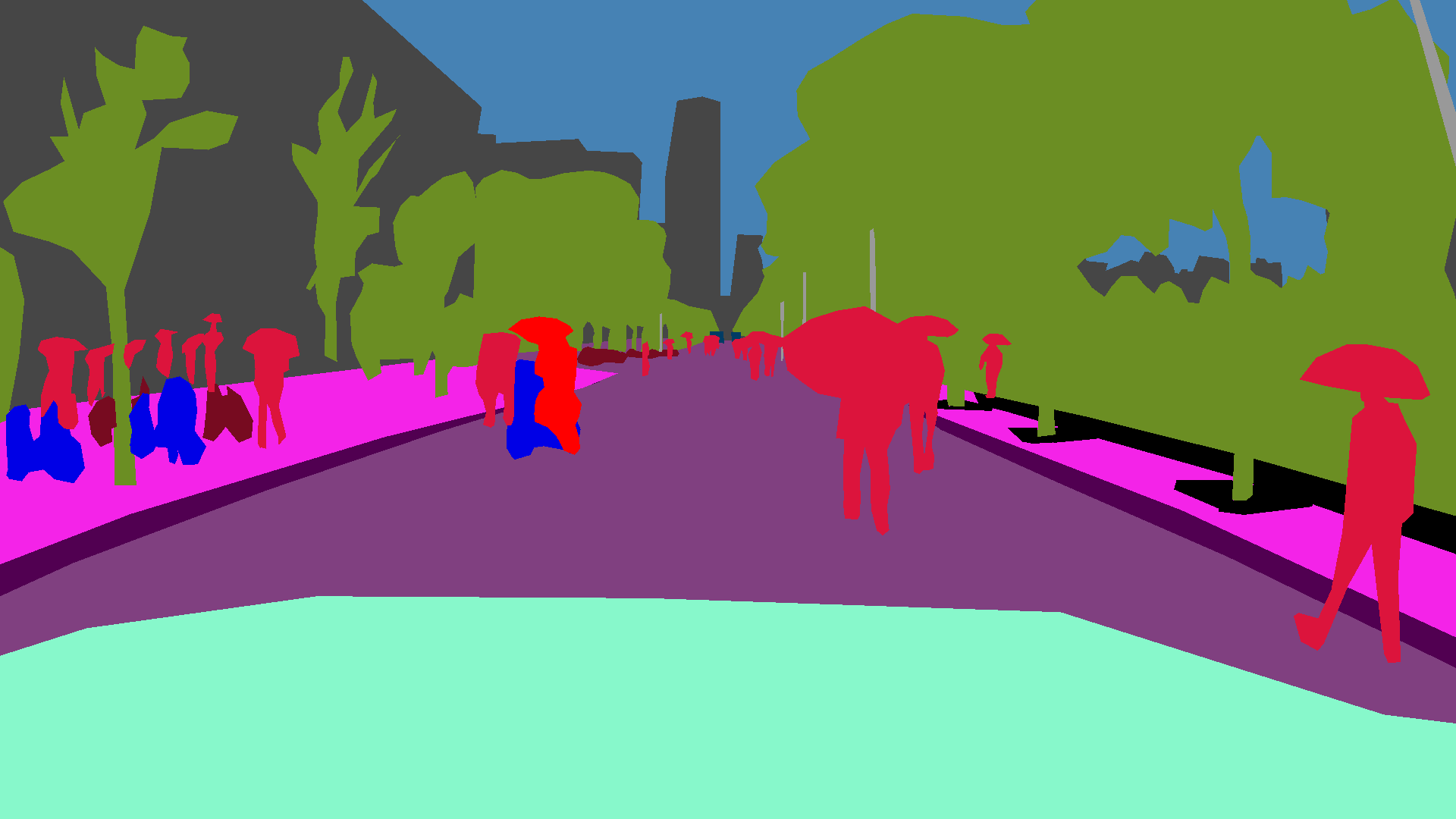}}
\hfill
\subfloat{\includegraphics[width=0.16\linewidth]{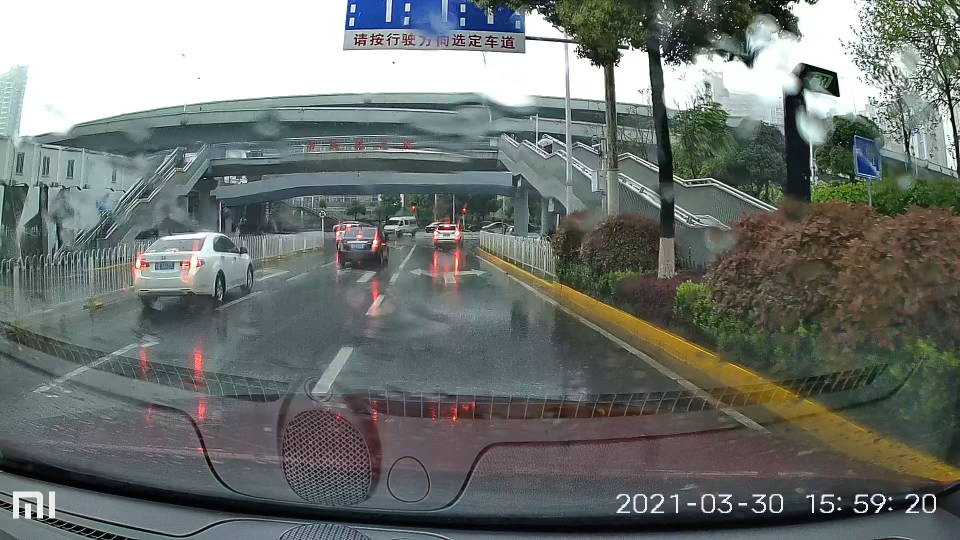}}
\hfill
\subfloat{\includegraphics[width=0.16\linewidth]{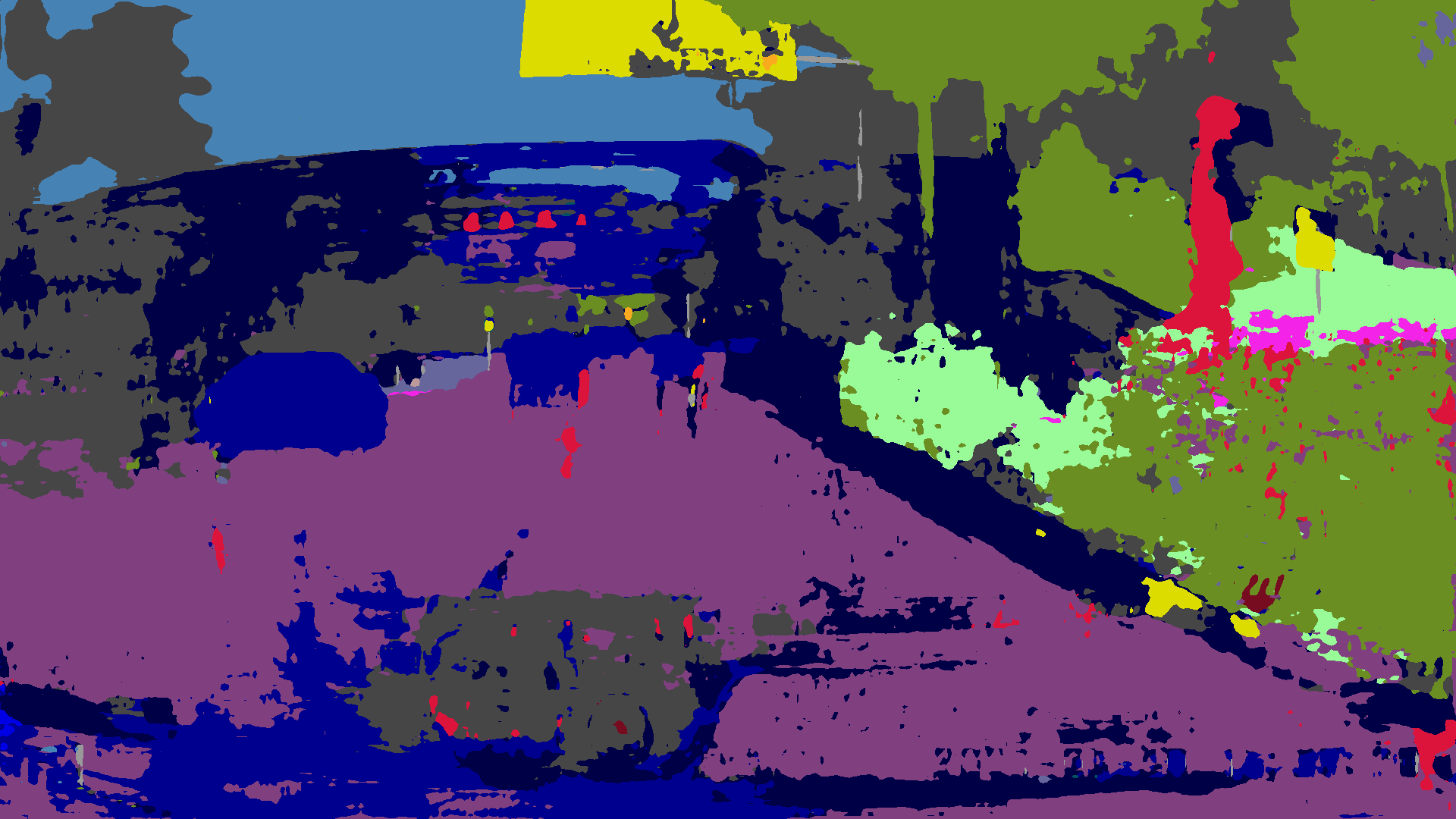}}
\hfill
\subfloat{\includegraphics[width=0.16\linewidth]{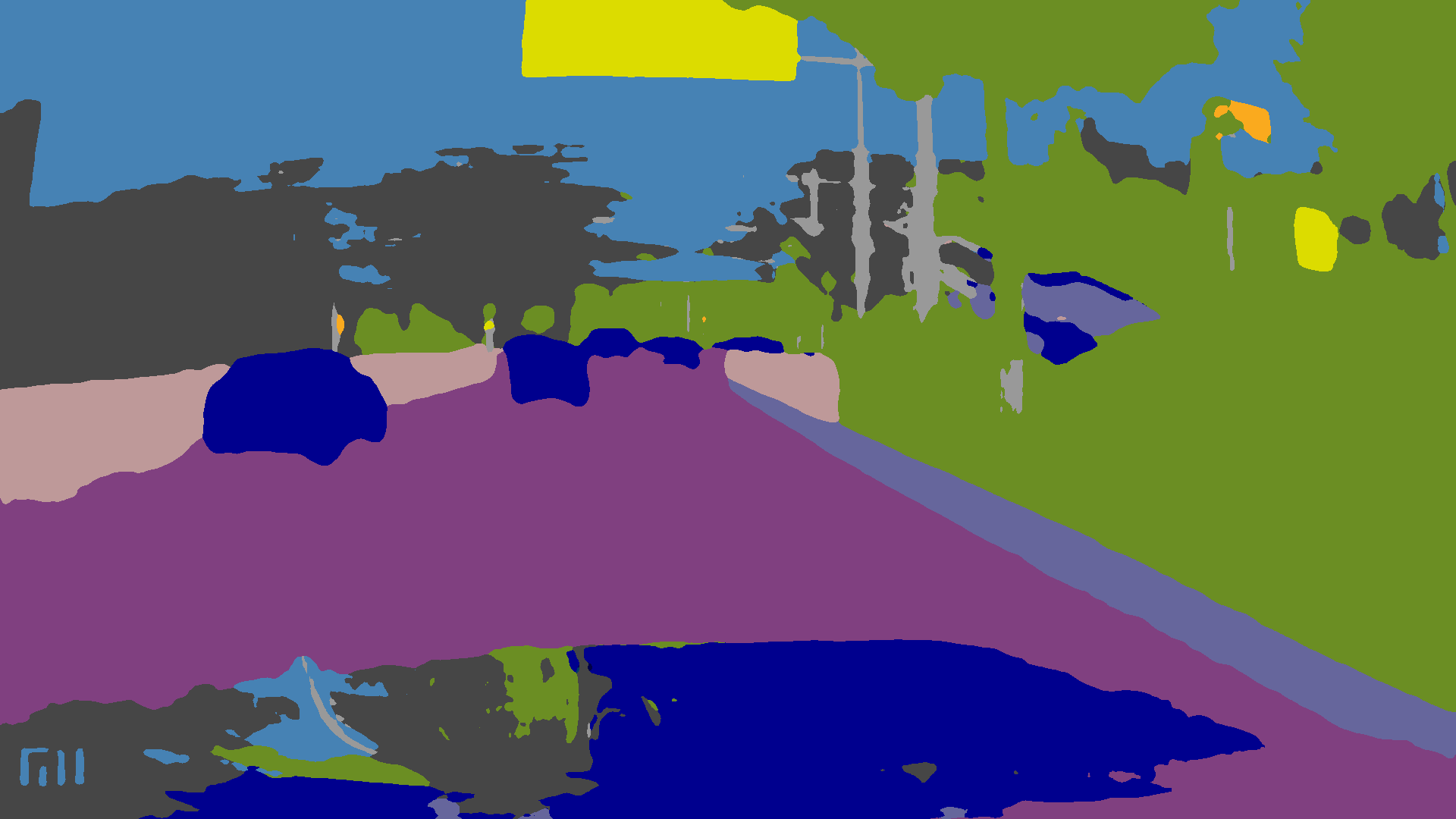}}
\hfill
\subfloat{\includegraphics[width=0.16\linewidth]{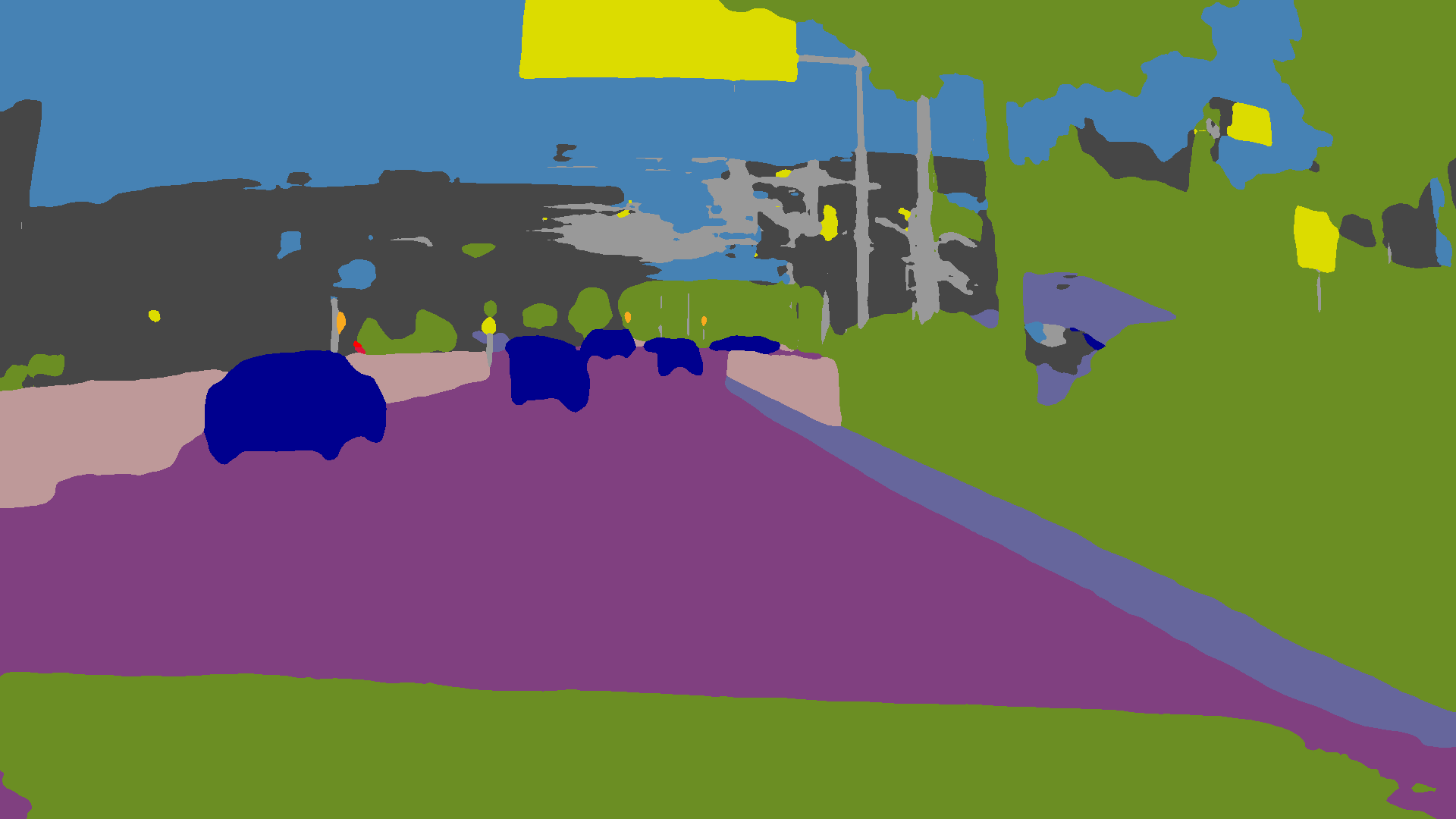}}
\hfill
\subfloat{\includegraphics[width=0.16\linewidth]{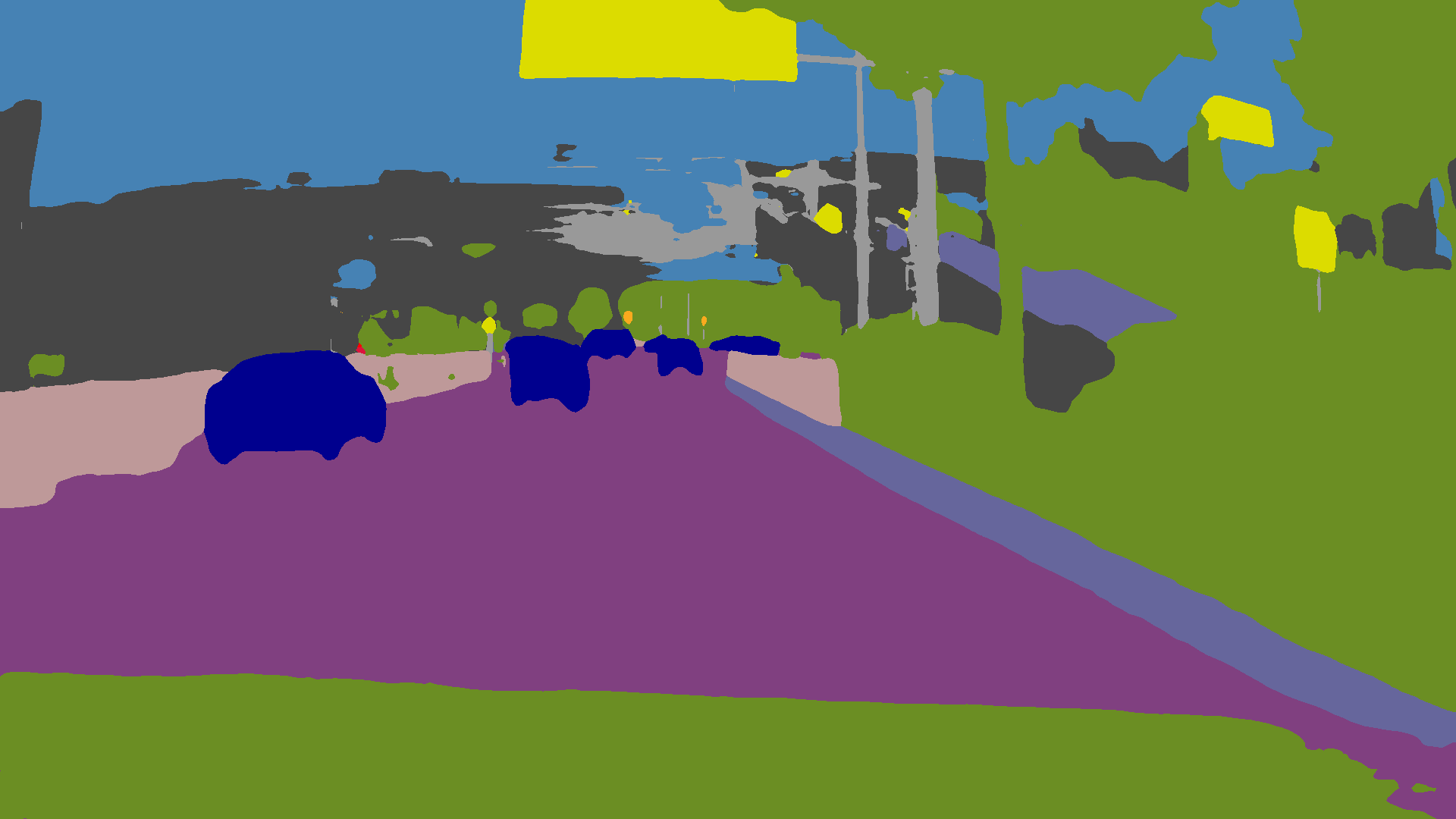}}
\hfill
\subfloat{\includegraphics[width=0.16\linewidth]{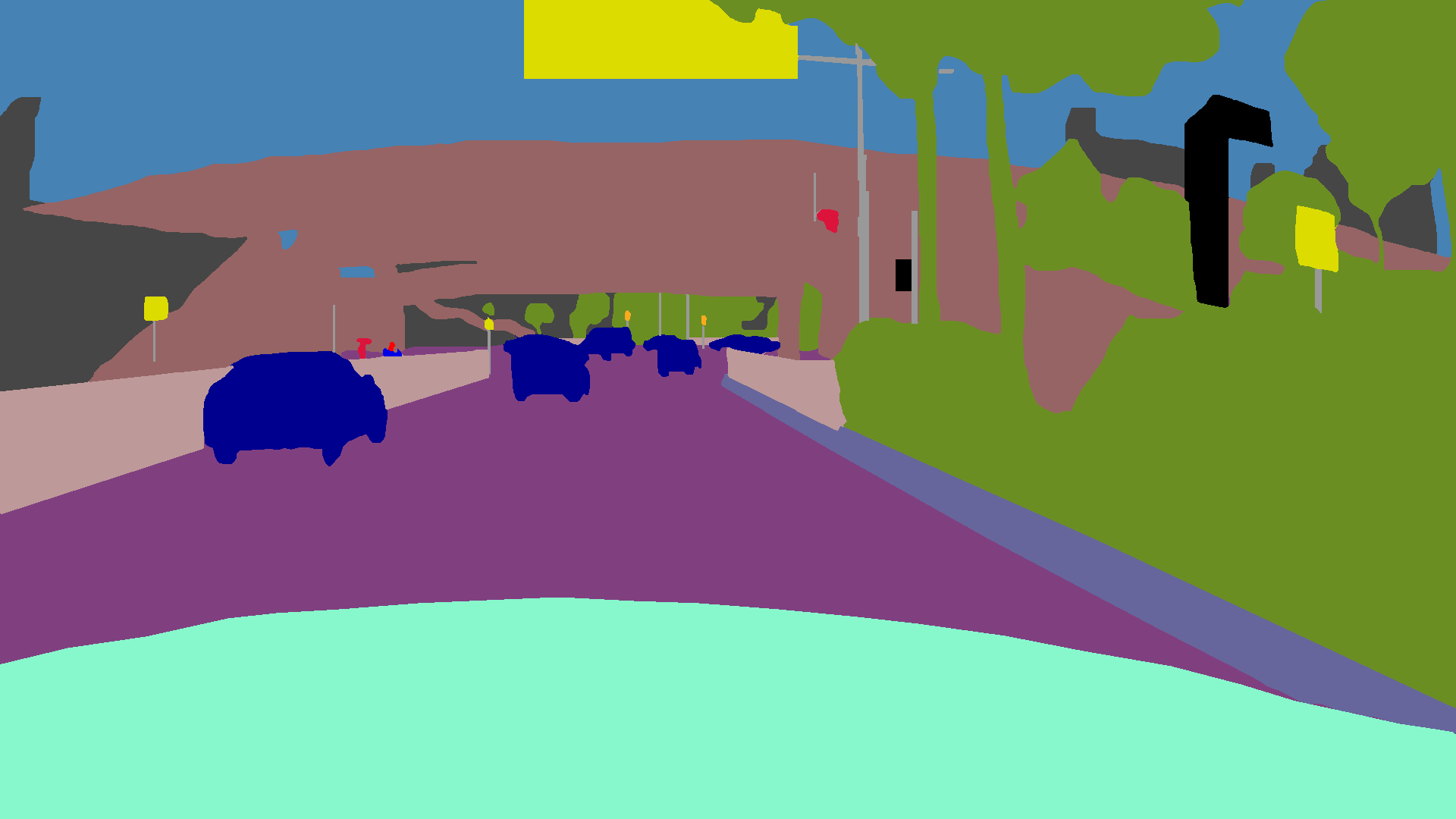}}
\hfill
\setcounter{subfigure}{0}
\subfloat[Rainy Scene]{\includegraphics[width=0.16\linewidth]{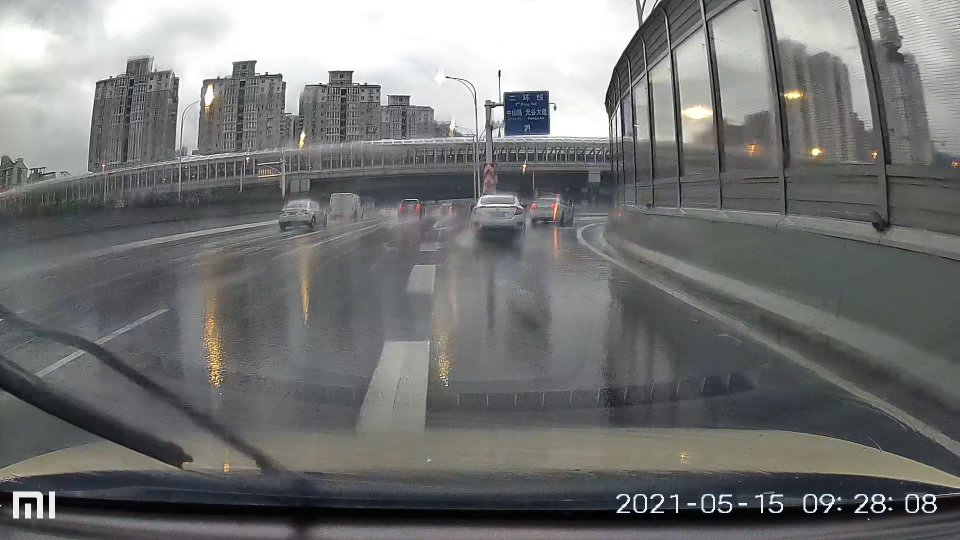}}
\hfill
\subfloat[Pre-trained]{\includegraphics[width=0.16\linewidth]{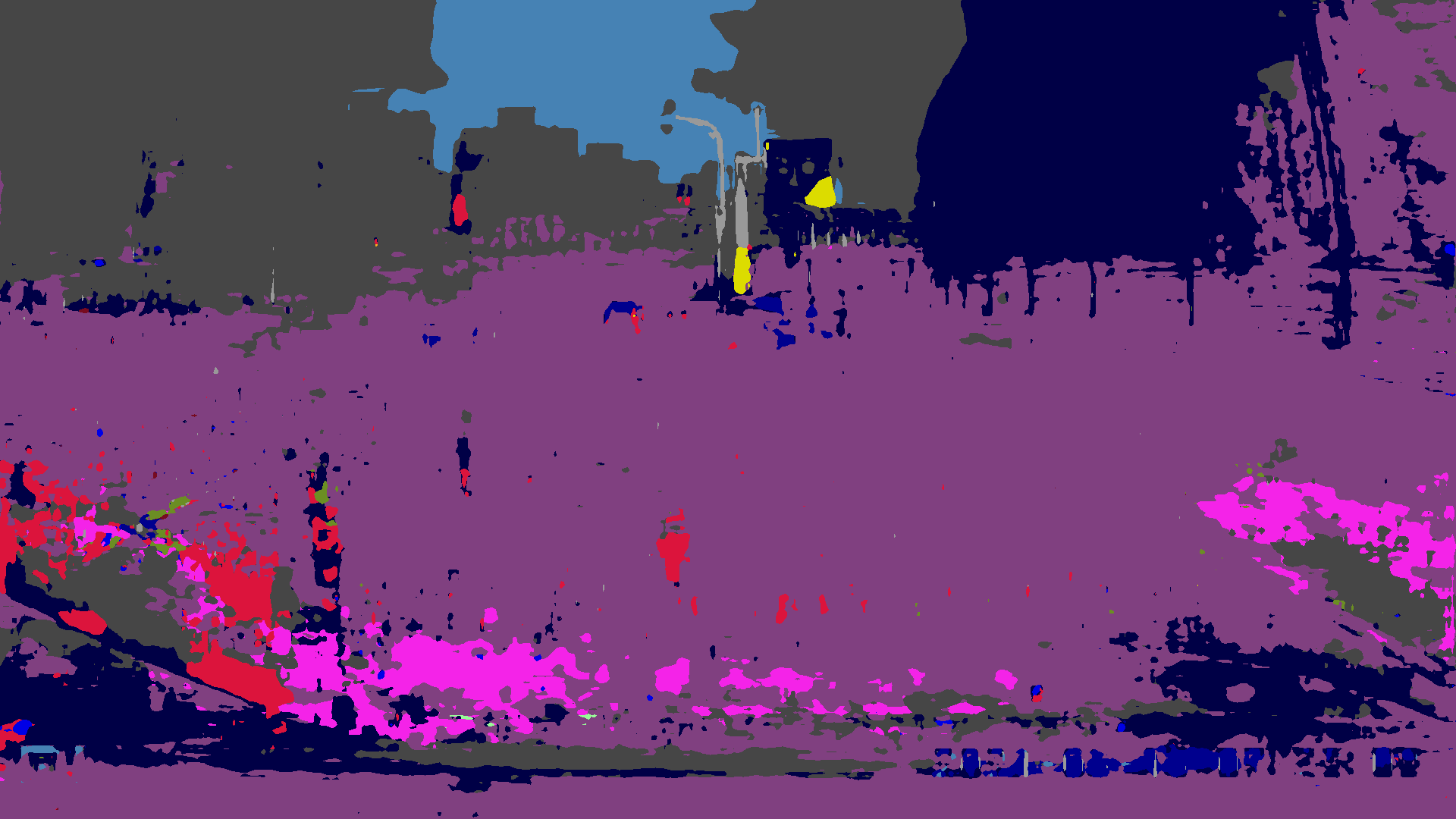}}
\hfill
\subfloat[Supervised]{\includegraphics[width=0.16\linewidth]{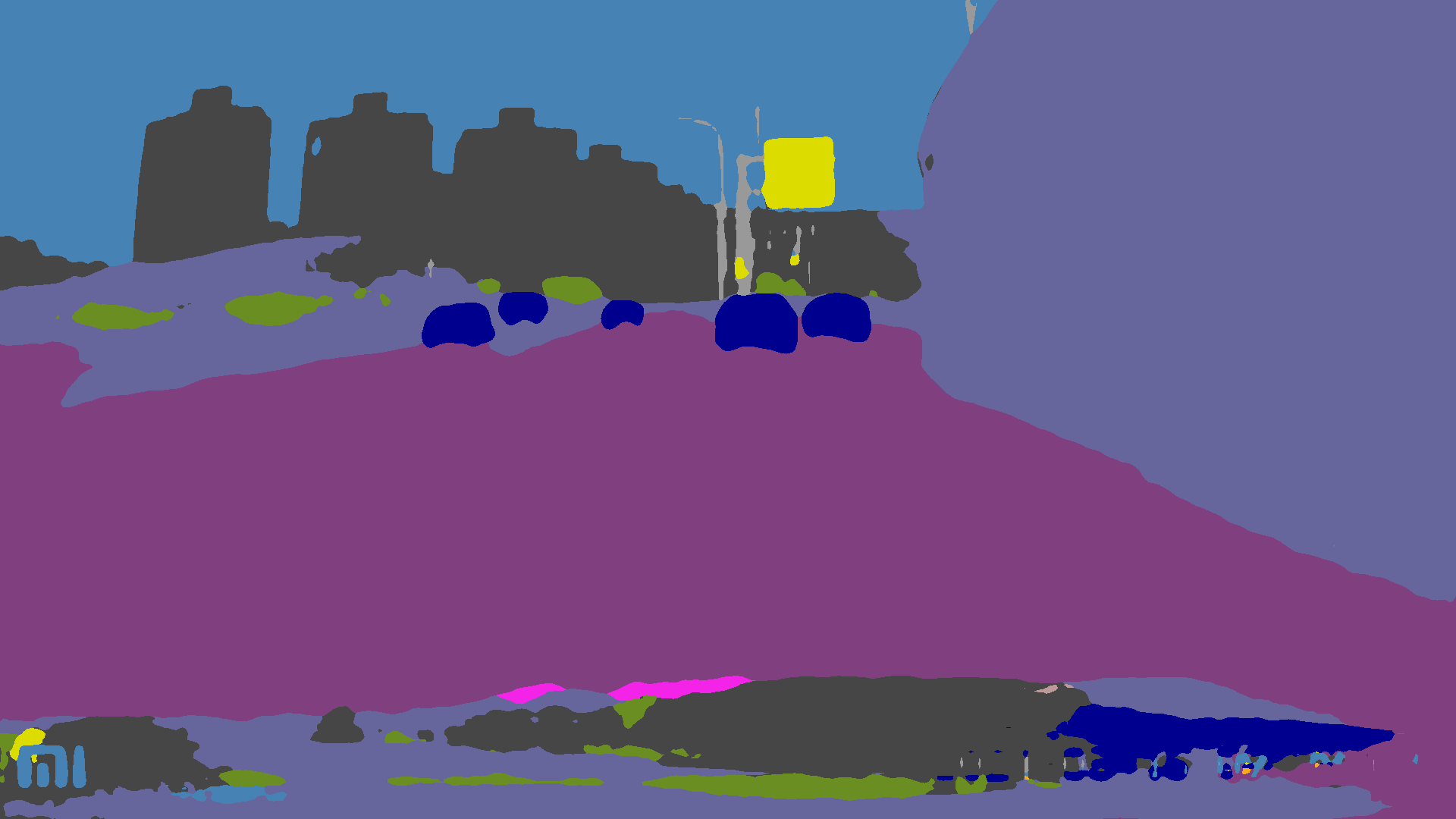}}
\hfill
\subfloat[Semi-supervised]{\includegraphics[width=0.16\linewidth]{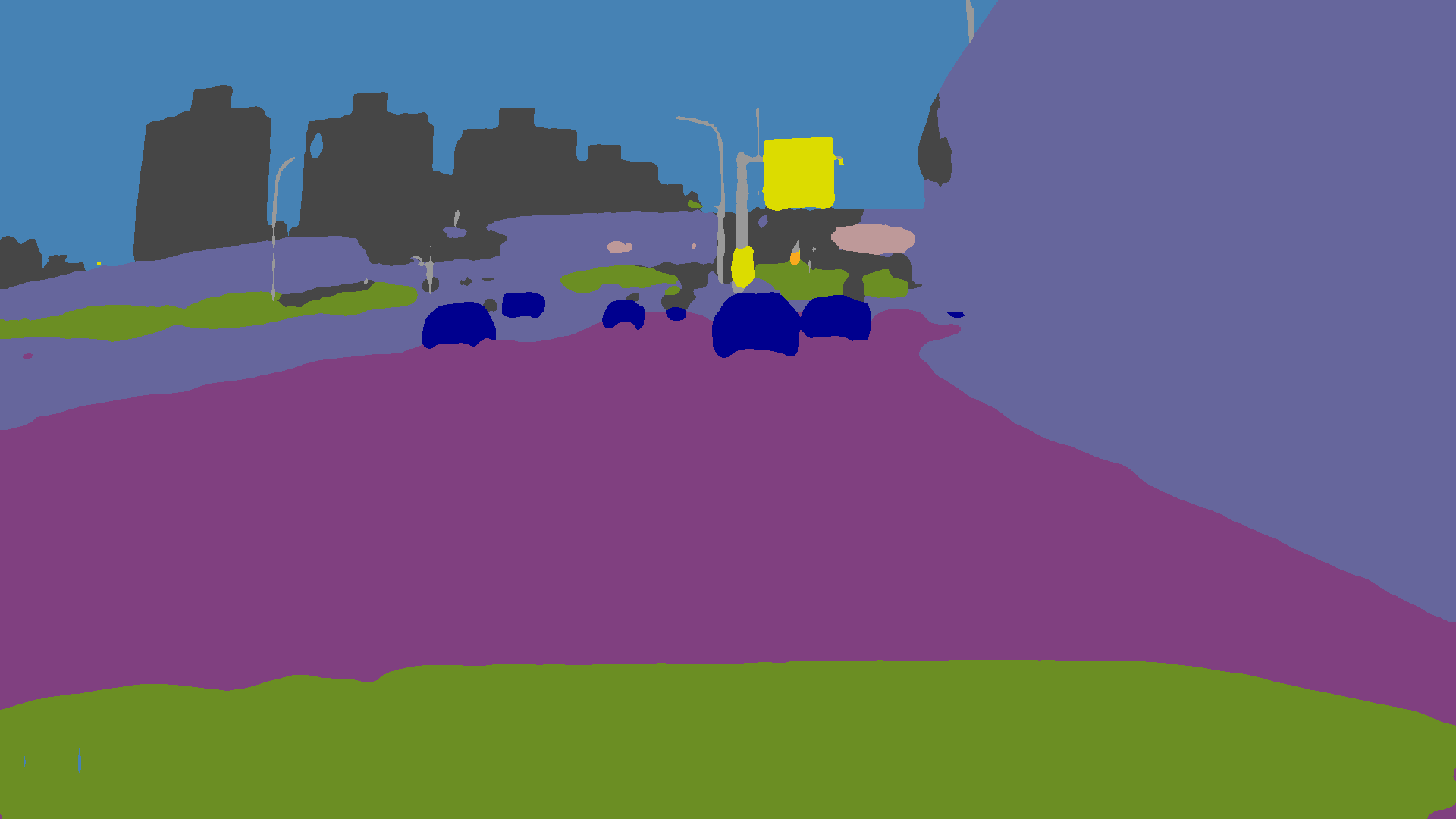}}
\hfill
\subfloat[Ours]{\includegraphics[width=0.16\linewidth]{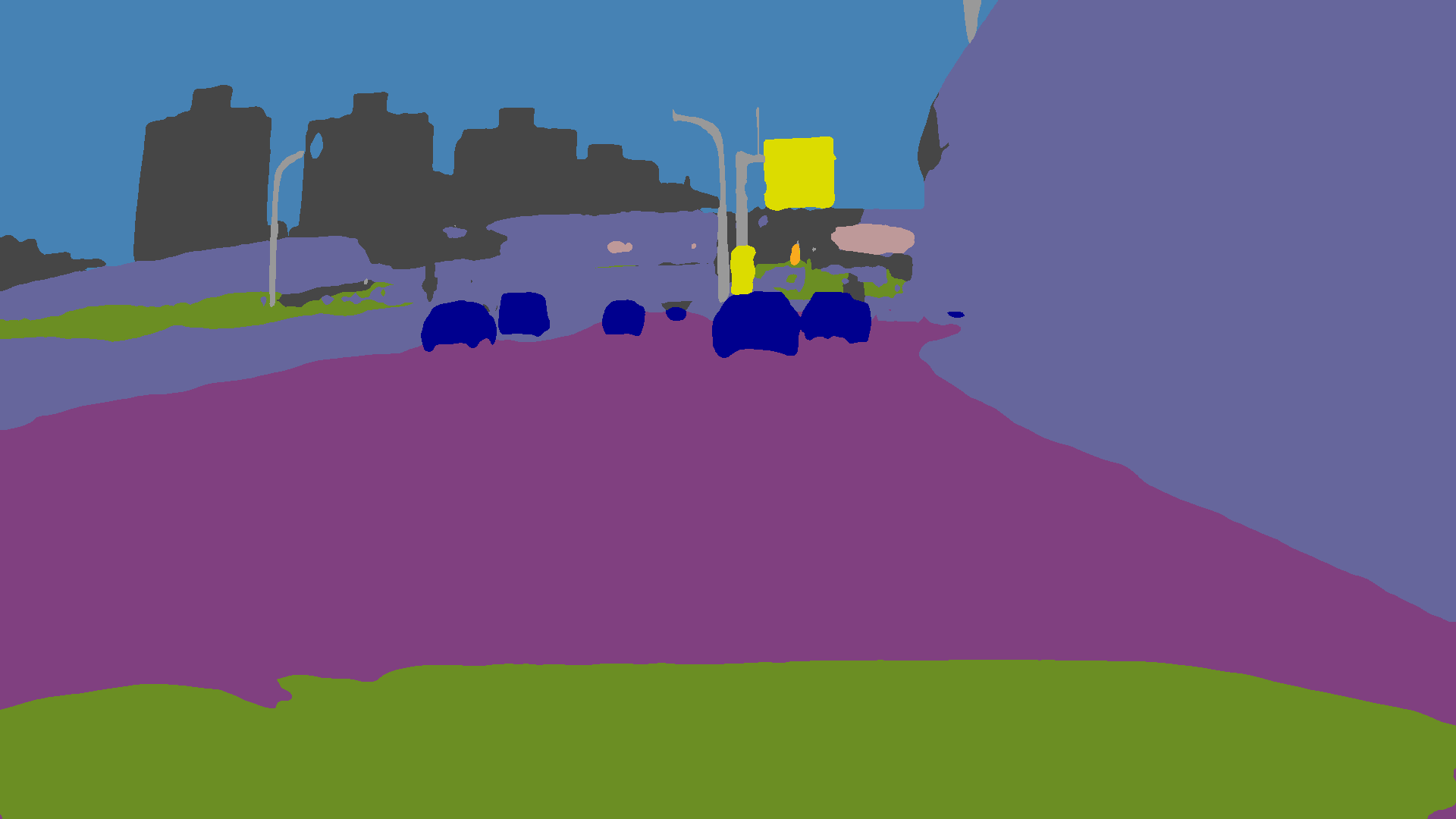}}
\hfill
\subfloat[Ground Truth]{\includegraphics[width=0.16\linewidth]{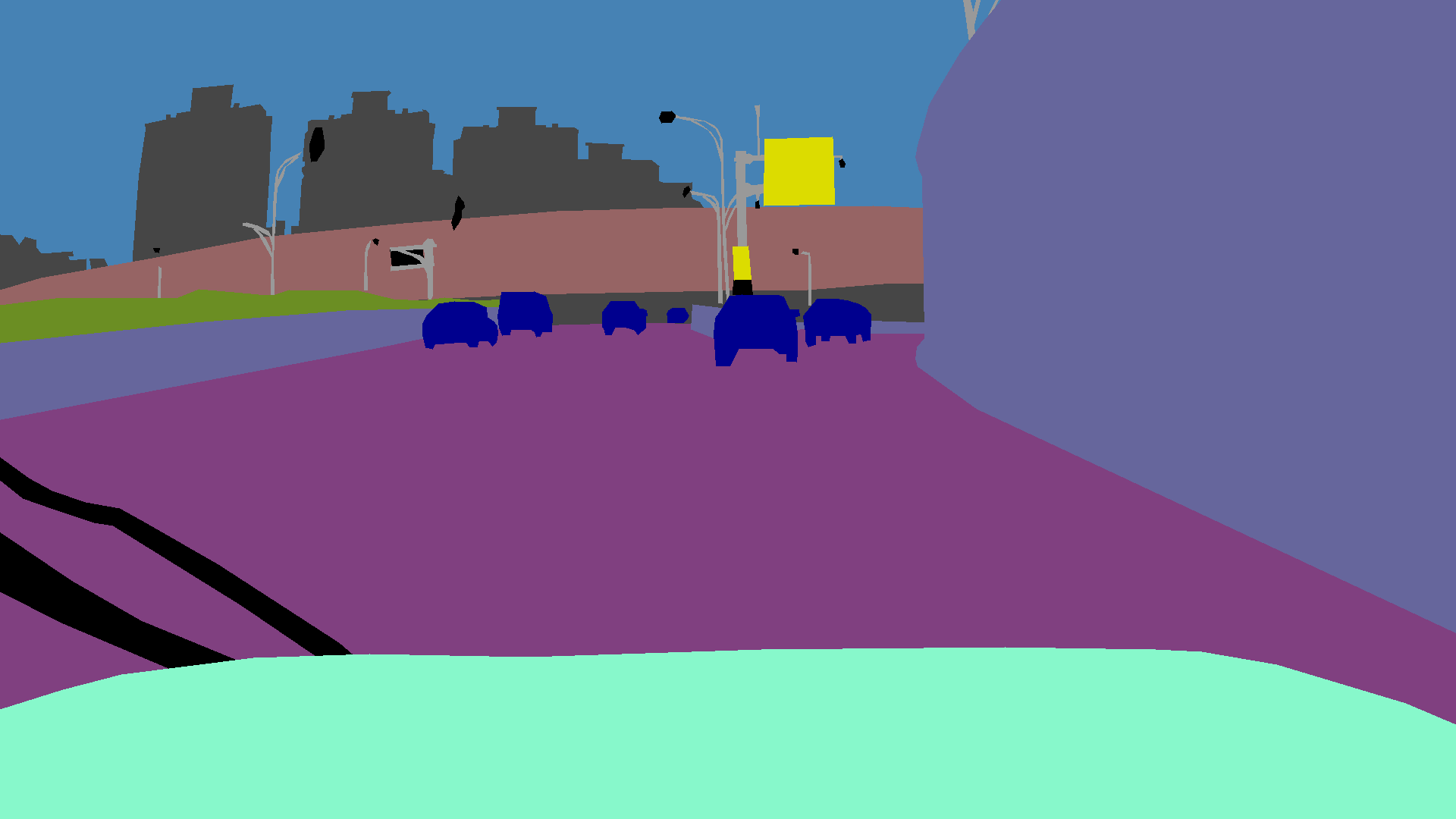}}
\caption{Visualization of segmentation results. Sub-figure (b) indicates the results obtained by a pre-trained model on cityscapes, (c) shows the results of a supervised model based on OHEM, (d) shows the results of the semi-supervised model in our framework, (e) displays the final segmentation results achieved by our framework, and (f) shows the ground truth. All of methods are based on U$^2$PL for conducting fair comparisons. }
\label{fig:vis}
\end{figure*}

Table~\ref{tab:1} presents the evaluation results of the top 5 teams on the Grand Challenge Proposal of the IEEE International Conference on Multimedia and Expo 2023: Seeing Through the Rain (STRAIN) - Track 1 - Semantic Segmentation under Real Rain Scene. The table displays the performance of each category, measured using the Intersection over Union (IoU) metric, as well as the overall performance for all categories, measured using the mean Intersection over Union (mIoU) metric. The results indicate that our framework has achieved state-of-the-art (SOTA) results. The ground truth of test dataset only released after finishing evaluation. Notably, the IoU value of our framework for the "person" class far exceed those of the other methods. This is mainly because SAM provide informative segmentation masks. As shown in Fig. \ref{fig:mask1}, SAM has strong segmentation capability for "person", and this advantage has been passed on to our framework. In addition, Fig. \ref{fig:vis} visualizes part of segmentation results on test dataset.

\subsection{Ablation Study} 

Table~\ref{tab:2} presents the results of our ablation study, which was conducted using the ground truth provided by the official test set. Specifically, the use of SAM for enhancing the segmentation results without filtering and sorting is indicated by "ENHANCE". Additionally, the use of filtering and sorting to process segmented entities is denoted by "w/ filter" and "w/ sort", respectively, which can be found on line 8 and 13 in Algorithm~\ref{alg:1}. The results in Table~\ref{tab:2} demonstrate the effectiveness of our proposed components. Moreover, we also visualize the segmentation masks generated by SAM with anchor-based prompting in Fig.~\ref{fig:mask} and make comparisons. Fig.~\ref{fig:mask1} shows that anchor-based prompting can capture small and difficult-to-segment entities.

\begin{table}[h]
\centering
\caption{The experimental results of ablation study}
\label{tab:2}
\begin{tabular}{cccccc}
\toprule
\textbf{Method} & \textbf{ENHANCE} & \textbf{w/ filter} & \textbf{w/ sort} & \textbf{mIoU}\\
\midrule
\multirow{4}{*}{Ours}            &            &            &            & 63.11\\
                                 & \checkmark &            &            & 62.72\\
                                 & \checkmark & \checkmark &            & 62.83\\
                                 & \checkmark & \checkmark & \checkmark & \textbf{64.24}\\
\bottomrule
\end{tabular}
\end{table}

\section{Conclusion}

In conclusion, this paper presents an innovative framework for addressing the challenges of semantic segmentation in rainy scenes. Our approach successfully harnesses the power of pre-trained foundation models without retraining by utilizing the entropy-based anchors generated by a semi-supervised base model. Our experiments, conducted on the Rainy WCity dataset, demonstrate that the proposed framework effectively leverages the pre-trained segmentation foundation model, leading to superior segmentation accuracy. Furthermore, our framework also provides insights and inspiration for active prompting in promptable foundation models.

However, there are still limitations to our framework. The accuracy of refinement essentially depends on the correctness of the guidance information provided by the semi-supervised base model, and our method relies on an additional model during inference, which limits its applicability to certain scenarios. In future work, we plan to investigate how to transfer knowledge from pre-trained segmentation foundation models to models in specific domains.

\bibliographystyle{IEEEtran}
\bibliography{conference_101719}

\end{document}